\documentclass[10pt,twocolumn,letterpaper]{article}

\usepackage{cvpr}
\usepackage{times}
\usepackage{epsfig}
\usepackage{graphicx}
\usepackage{amsmath}
\usepackage{amssymb}
\usepackage{multirow}

\usepackage[breaklinks=true,bookmarks=false]{hyperref}

\cvprfinalcopy 

\def\cvprPaperID{****} 

\begin{document}

\title{RUC+CMU: System Report for Dense Captioning Events in Videos}

\author{Shizhe Chen\textsuperscript{1}, Yuqing Song\textsuperscript{1}, Yida Zhao\textsuperscript{1}, Jiarong Qiu\textsuperscript{1}, Qin Jin\textsuperscript{1\thanks{Corresponding author.}} and Alexander Hauptmann\textsuperscript{2}\\
\textsuperscript{1} Renmin University of China, \textsuperscript{2} Carnegie Mellon University\\
{\tt\small \{cszhe1, syuqing, zyiday, jiarong\_qiu, qjin\}@ruc.edu.cn, alex@cs.cmu.edu}
}

\maketitle

\begin{abstract}
This notebook paper presents our system in the ActivityNet Dense Captioning in Video task (task 3).
Temporal proposal generation and caption generation are both important to the dense captioning task.
Therefore, we propose a proposal ranking model to employ a set of effective feature representations for proposal generation, and ensemble a series of caption models enhanced with context information to generate captions robustly on predicted proposals.
Our approach achieves the state-of-the-art performance on the dense video captioning task with 8.529 METEOR score on the challenge testing set.

\end{abstract}

\section{Task Introduction}

Most natural videos contain multiple events.
Instead of generating a single sentence to describe the overall video content,
the dense video captioning task aims to localize the event and generate a series of sentence to describe each event.
This task is more challenging than the single sentence video captioning task, which requires to generate good temporal event proposals, consider the correlations of different events in the video and so on.
\section{Proposed Approach}

\begin{figure*}
	\begin{center}
		\includegraphics[width=1\linewidth]{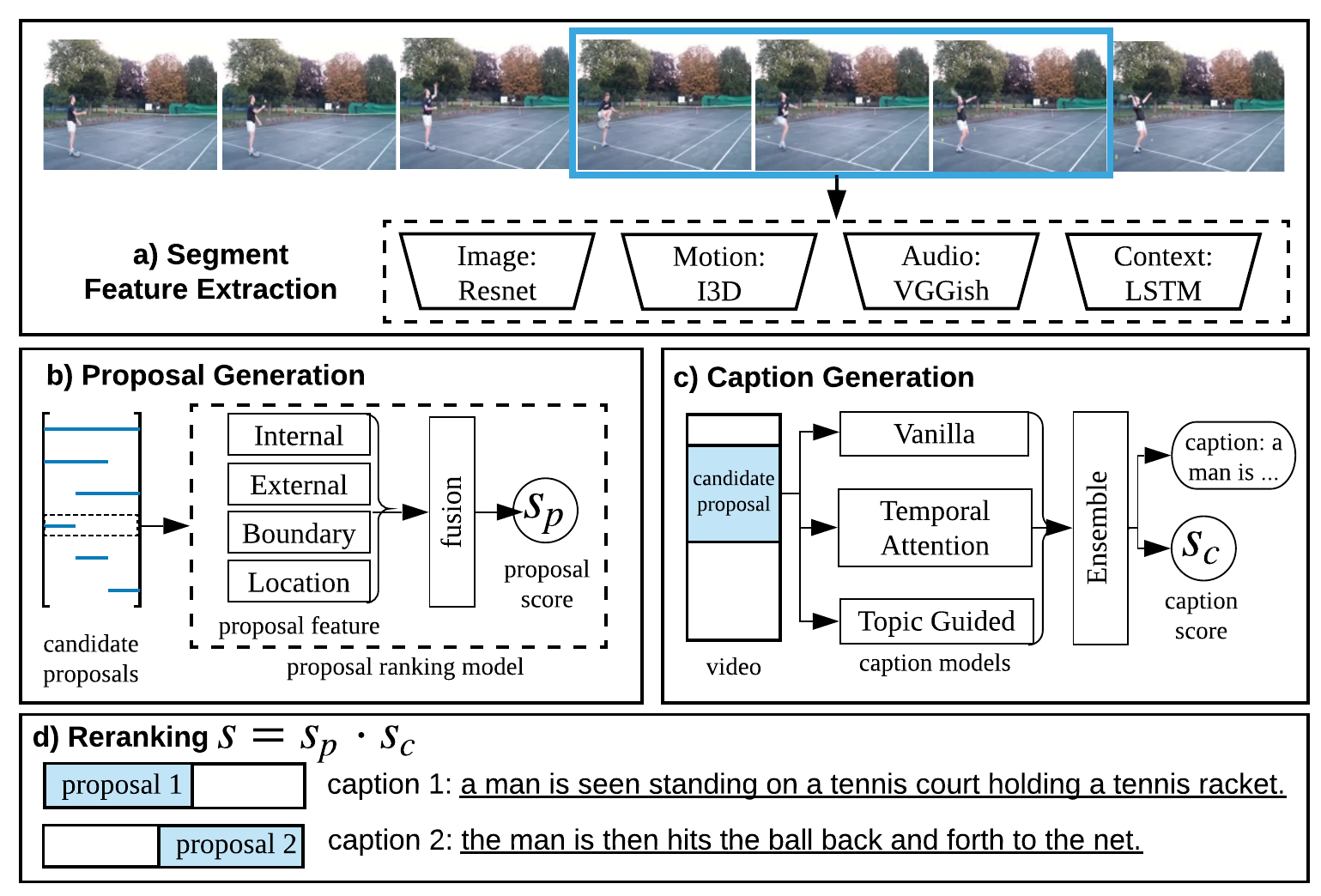}
	\end{center}
	\caption{Framework of our proposal approach, which consists of four components: 1) segment feature extraction to transfer the video into a sequence of multimodal features; 2) proposal generation which contains a proposal ranking model to select good event proposals; 3) caption generation which employs various caption models to generate accurate event descriptions; and 4) re-ranking to select event captions with both high proposal and caption score.}
	\label{fig:framework}
\end{figure*}

The framework of our approach is presented in Figure~\ref{fig:framework}, which consists of four components: 1) segment feature extraction; 2) proposal generation; 2) caption generation; and 4) re-ranking.
In this section, we introduce each component of the framework in details.

\subsection{Segment Feature Extraction}
We divide the video clip into non-overlapping segments and extract features for each segment. 
The length of the segment is set to be 64 frames in our work.
Since the video contains multi-modal information, we first extract three types of deep features from different modalities, which are: 1) image modality: Resnet features \cite{he2016deep} pretrained on the ImageNet dataset; 2) motion modality: I3D features \cite{carreira2017quo} pretrained on the Kinetics dataset; and 3) audio modality: VGGish features \cite{hershey2017cnn} pretrained on the Youtube8M dataset.

As shown in previous works \cite{krishna2017dense}, the context information plays an important role in generating proper captions for an event proposal.
Therefore, we utilize a bidirectional LSTM to capture the context information and extract the hidden states of LSTM as our context feature.
The LSTM employs the aforementioned three types of deep features as input, and is trained to predict concepts in groundtruth captions in each step.
In such a way, the LSTM learns the bidirectional context for each segment to generate captions.

After the feature extraction, the video is represented as a sequence of segment-level features.

\subsection{Proposal Generation}
We adopt a two-stage pipeline to generate temporal proposals.
Firstly, a heuristic sliding window method is exploited to generate a series of candidate proposals for each video.
Then, we train a proposal ranking model to select proposals that are of high tiou (temporal intersection over union) with groundtruth proposals.

\textbf{Candidate Proposal Generation}

In order to generate candidate proposals with high recalls, we apply the sliding window approach on the video clip.
Assuming $w$ is the length of the window, we slide the window over the clip with the shift of $w/4$.
The window lengths are generated according to lthe length distribution of groudtruth proposals and the length of the video.
We first cluster proportions of the groundtruth proposal in the video  into $K$ centers $\{w^p_1, \cdots, w^p_K  \}$.
Then we set the window lengths for each video to be $w_k=w^p_k \cdot l$ for $k=1, \cdots, K$, where $l$ is the length of the video.

\textbf{Proposal Ranking Model}

The proposal ranking model is trained to filter out inappropriate candidate proposals.
We consider a good temporal proposal to satisfy the following conditions: 
1) the event in the proposal is meaningful;
2) the event in the proposal is different from its context;
3) the boundaries of the proposal contain variance;
and 4) the location of the proposal is  satisfied with groundtruth distributions.
Therefore, we propose four different features to satisfy the conditions above:
1) internal feature: mean pooling of segment features in the proposal to represent events in the proposal;
2) external feature: mean pooling of segment features in the context to represent contextual events;
3) boundary feature: the difference of the segment feature near the proposal boundary to represent the boundary variance;
and 4) the proportion of the location and duration of the proposal.
We utilize a two-layers feed-forward neural network to fuse these features and predict the proposal score $s_p$ of the proposal.
During training, candidate proposals with tiou above 0.7 are as positive samples and tiou less than 0.5 are as negative samples.

\subsection{Caption Generation}
In order to generate accurate and diverse video captions, we employ three different caption models and ensemble them to generate the caption for each event proposal.

\textbf{Vanilla Caption Model} \cite{jin2017knowing} is the baseline model for the video captioning task.
It consists of a multimodal video encoder and a LSTM language decoder.
Since the context is vital to generate consistent captions for the proposal, we enhance the encoder with the LSTM context features \cite{wang2018bidirectional}.

\textbf{Temporal Attention Caption Model} \cite{yao2015describing} improves over the vanilla caption model via paying attention to relevant segments in the video to generate each word.
To incorporate the context, we also enhance the encoder in the attention model with the contexts of the boundaries. 

\textbf{Topic Guided Caption Model} \cite{chen2017generating} utilizes the video topics to guide the caption model to generate topic-aware captions.
Since there are 200 manual labeled categories in the ActivityNet dataset, we directly use these categories as our topics.
We train a topic predictor which is a single-layer feed-forward neural network to predict the category probabilities of each proposal.
As the size of the dataset is not large, we adopt the Topic Concatenation in Decoder version in \cite{chen2017generating} to guide the caption generation which requires fewer parameters than TGM in \cite{chen2017generating}.

We firstly use the cross entropy loss to pretrain all the caption models, which optimizes the likelihood of the groundtruth captions.
But such training approach  suffers from the exposure bias and evaluation mismatch problems.
Therefore, we employ the self-critical reinforcement learning \cite{rennie2017self} to further train our caption models, which is the state-of-the-art approach in image captioning and alleviates the above two problems.
CIDEr and METEOR are weighted as our reinforcement reward.

\textbf{Caption Model Ensemble} aims to make use of various caption models. We ensemble the word prediction of each model at every step. Beam search with beam size of 5 is used to generate the final caption with probability score $s_c$.

\subsection{Re-ranking}
Since both the proposal quality and the caption quality influence the evaluation of dense captions, we re-rank the captions of different proposals by $s=s_p \cdot s_c$. The top 10 captions with their proposal are selected.

\section{Experimental Results}

\subsection{Experimental Settings}
\textbf{Dataset:} The ActivityNet Dense Caption dataset \cite{krishna2017dense} is used in our work. We follow the official split with 10,009 videos for training, 4,917 videos for validation and the remaining 5,044 videos for testing. The groundtruth of the testing videos are unknown.
For the final submission, we enlarge our training set with part of validation set to future improve the performance, which  contains 14,009 videos for training and 917 videos for validation.

\textbf{Evaluation Metrics:} 
We employ the precision and recalls to evaluate the performance of proposals.
To evaluate the captions, we first evaluate the performance of the caption using the groundtruth proposal.
And then we use the same metric as \cite{krishna2017dense} to evaluate the captions of predicted proposals, which computes the caption performance for proposals with tiou 0.3, 0.5 and 0.7 with the groundtruth.

\subsection{Evaluation of Proposals}

Table~\ref{tab:proposal} presents the performance of our proposal generation approach.
For the sliding window candidate proposal generation, we use 20 clusters to generate sliding window, which leads to 241 proposals for each video.
We can see that the heuristic sliding window approach achieves remarkable recall (0.98 on average), while the precision of the proposal is quite low.
After applying the proposal ranking model, we select proposals that contain proposal score $s_p > 0.5$ which results in 53 proposals on average for each video. 
The precision is significantly improved (0.71 vs 0.28) with minor recall decrease, which demonstrates the effectiveness of our proposal ranking model.

\begin{table}
	\centering
	\caption{Performance of the proposal generation approach. P and R are short for precision and recall.}
	\label{tab:proposal}
	\begin{tabular}{c|c|c|ccc|c} \hline
		& \#props & metric & 0.3 & 0.5 & 0.7 & avg \\ \hline
		\multirow{2}{*}{\begin{tabular}[c]{@{}c@{}}sliding\\ window\end{tabular}} & \multirow{2}{*}{241} & P & 0.45 & 0.27 & 0.12 & 0.28 \\
		&  & R & 0.99 & 0.99 & 0.95 & 0.98 \\ \hline
		\multirow{2}{*}{\begin{tabular}[c]{@{}c@{}}proposal \\ ranking\end{tabular}} & \multirow{2}{*}{53} & P & 0.97 & 0.77 & 0.38 & 0.71 \\
		&  & R & 0.91 & 0.85 & 0.76 & 0.84 \\ \hline
	\end{tabular}
\end{table}

\subsection{Evaluation of Captions}
Table~\ref{tab:caption_prop} shows the caption performance using groundtruth proposals.
We can see that the performance of different models are competitive with each other, and the ensemble of these models achieves the best performance consistently on different caption metrics.
For the predicted proposals, the performance is dropped a little due to the imperfect proposal, which shows the robustness of our caption model on imperfect proposals.
The significant decrease of CIDEr score mainly results from the more proposals in the predicted version than the groundtruth, which makes the tf-idf statistics different.

\begin{table}
	\centering
	\caption{Performance of difference caption models.}
	\label{tab:caption_prop}
	\begin{tabular}{c|c|ccc} \hline
proposal &	model & Bleu4 & Meteor & CIDEr \\ \hline
	\multirow{4}{*}{groundtruth} & vanilla & 3.62 & 13.37 & 52.36 \\
	&	attention & 3.69 & 13.21 & 53.45 \\
	&	topic guided & 3.46 & 13.71 & 51.53 \\ \cline{2-5}
	&	ensemble & \textbf{3.97} & \textbf{13.75} & \textbf{56.45} \\ \hline
predicted	&	ensemble & 4.00 & 12.44 & 31.10  \\ \hline
	\end{tabular}
\end{table}

\subsection{Submission}
For the final submission, we train our caption models on the bigger training set and utilize the smaller validation set to select models.
The performance of the submitted model is presented in Table~\ref{tab:submission}.
More training data brings small improvement, and our model achieves 8.529 METEOR score on the testing set.

\begin{table}
	\centering
	\caption{Performance of the submitted models.}
	\label{tab:submission}
	\begin{tabular}{c|ccc} \hline
		& Bleu4 & Meteor & CIDEr \\ \hline
		val\_small & 3.92 & 12.67 & 31.92 \\
		testing & - & 8.529 & - \\ \hline
	\end{tabular}
\end{table}

\section{Conclusion}
In this work, we propose a system with four components to generate dense captions in videos, which achieves significant improvements on the dense video captioning task.
Our results show that it is important to utilize context-related features for both the proposal generation and caption generation. 
In the future, we will explore to unify the system in an end-to-end way to improve the proposal module with captions and generate more diverse caption for the events.

{\small
\bibliographystyle{ieee}
\bibliography{egbib}
}

\end{document}